\documentclass[letterpaper, 10 pt, conference]{ieeeconf}  

\IEEEoverridecommandlockouts                              

\usepackage{times}
\usepackage{soul}
\usepackage{url}
\usepackage{graphicx}
\usepackage{amsmath}

\usepackage{amsthm}
\usepackage{booktabs}
\usepackage{algorithm}
\usepackage{algorithmic}
\usepackage{algorithm}
\usepackage{algorithmic}
\usepackage{amsfonts}  
\usepackage{subfigure}
\usepackage{amsmath}
\usepackage{rotating}
\usepackage{multirow}
\usepackage{pgffor}
\usepackage{lineno}
\usepackage{wrapfig}

\newcommand{\bs}{\mathbf{s}}
\newcommand{\ba}{\mathbf{a}}

\title{\LARGE \bf
A Real-World Quadrupedal Locomotion Benchmark for Offline Reinforcement Learning 
}

\author{Hongyin Zhang, Shuyu Yang, Donglin Wang*
\thanks{*Corresponding author.}
\thanks{All the authors are with Machine Intelligence Lab (MiLAB), School of Engineering, Westlake University, Hangzhou 310024, China, and Institute of Advanced Technology, Westlake Institute for Advanced Study, Hangzhou
310024, China.}%
}

\begin{document}

\maketitle
\thispagestyle{empty}
\pagestyle{empty}

\begin{abstract}
Online reinforcement learning (RL) methods are often data-inefficient or unreliable, making them difficult to train on real robotic hardware, especially quadruped robots.
Learning robotic tasks from pre-collected data is a promising direction.
Meanwhile, agile and stable legged robotic locomotion remains an open question in their general form.
Offline reinforcement learning (ORL) has the potential to make breakthroughs in this challenging field, but its current bottleneck lies in the lack of diverse datasets for challenging realistic tasks.
To facilitate the development of ORL, we benchmarked 11 ORL algorithms in the realistic quadrupedal locomotion dataset. 
Such dataset is collected by the classic model predictive control (MPC) method, rather than the model-free online RL method commonly used by previous benchmarks.
Extensive experimental results show that the best-performing ORL algorithms can achieve competitive performance compared with the model-free RL, and even surpass it in some tasks. 
However, there is still a gap between the learning-based methods and MPC, especially in terms of stability and rapid adaptation.
Our proposed benchmark will serve as a development platform for testing and evaluating the performance of ORL algorithms in real-world legged locomotion tasks.
\end{abstract}

\section{Introduction}
Recently, as an integral part of artificial intelligence, deep reinforcement learning (DRL) has achieved breakthroughs in various application areas \cite{DBLP:journals/nature/MnihKSRVBGRFOPB15,DBLP:journals/nature/SilverHMGSDSAPL16,eth2020}.
DRL algorithms have indeed achieved great success in simulation, but are limited in practical applications. 
Complex systems such as autonomous vehicles and robots are pretty expensive to operate and difficult to accurately simulate.
Inappropriate policies can have unbearable consequences.
To alleviate this problem, offline reinforcement learning (ORL) \cite{DBLP:books/sp/12/LangeGR12,DBLP:journals/corr/abs-2005-01643} has recently gained increasing attention from the DRL community.
Under this paradigm, the agent learns an optimal policy to maximize cumulative reward directly from previously collected datasets without any interaction with the environment. 

Many factors such as non-standardized evaluation protocols, different datasets, and lack of algorithm baselines can make validating and evaluating ORL algorithms difficult. 
A direct consequence is that performance comparisons among algorithms become ambiguous and hard to tell apart. 
Several works utilized trained online DRL algorithms to collect data, but these benchmarks are not designed with ORL in mind \cite{DBLP:journals/corr/abs-1911-11361}. 
Recent works have established benchmarks for ORL research and proposed standard evaluation protocols. 
But these benchmarks are mainly based on simulation environments and are far from real applications \cite{DBLP:journals/corr/abs-2004-07219}.

Therefore, we expect to construct an evaluation benchmark composed entirely of real empirical data, which ideally should: a) be able to encompass tasks of varying difficulties and reflect the challenges of ORL algorithms in real-world tasks, b) enable more researchers in the ORL community to utilize and further refine the dataset, and c) promote the development of ORL algorithms in the realistic applications.
Specifically, the dataset we currently construct comes from the quadruped robot \textit{Unitree A1}\footnote{https://www.unitree.com/products/a1/}, which can complete walking, uphill, turning and other behaviors. 

There are several important reasons why real-world quadrupedal locomotion tasks are used to benchmark ORL algorithms.
First, ORL has developed rapidly in recent years, and has achieved excellent results in the simulation environment, but there is still a long way from the actual landing. 
Second, DRL-based quadrupedal locomotion research is in its infancy for academia, and there is no public unified standard dataset yet. 
Compared with application platforms such as robotic arms and wheeled unmanned vehicles, the stable control of legged robots is more challenging, so it can better verify the effectiveness of the ORL algorithms.
Third, \textit{Unitree A1} is a low-cost quadruped robot that serves as a platform for deployment and verification of DRL algorithms.


On the other hand, previous ORL benchmarks mainly utilized model-free online RL to collect data.
For legged robot locomotion tasks, there have been some video demonstrations of the model-free online RL algorithms on the \textit{Unitree A1}, \textit{Go1} robot \cite{nahrendra2023dreamwaq,margolis2023walk}. 
The sim2real problem is also alleviated by domain randomization methods (terrain randomization, observational noise, motion delay, etc.).
However, the online RL policy deployed on quadrupedal robot is actually a sub-optimal control method, and its performance is not as good as classical model predictive control (MPC) \cite{DBLP:conf/iros/CarloWKBK18} method. 
More comparison and analysis in the experiments section.
Current online RL policies can certainly be used to collect real data, and ORL policies trained on that data may also be comparable in performance to online RL policies.
In fact, we think the quadrupedal tasks contained in such dataset are trivial for the ORL algorithms.
If current ORL algorithms can easily accomplish these tasks, it is difficult for the benchmark to bring more novel insights to further research.
Only when the dataset is challenging enough can the main flaws of the ORL algorithms be highlighted when faced with real-world tasks.
These deficiencies are the main problems that ORL urgently needs to solve in the face of practical application research.
Therefore, we adopt a relatively mature classical MPC method to collect data.

In this work, we construct a novel real-world quadrupedal locomotion benchmark.
To the best of our knowledge, it is the first ORL benchmark based on the realistic quadruped robot.
We focus on the design of tasks with different degrees of difficulty in typical real-world environments, which will help to examine the challenges that ORL algorithms may face in practical applications.
We benchmark 11 state-of-the-art ORL algorithms and provide reference implementations on real-world quadrupedal robot.
We also quantitatively analyzes the gap among learning-based methods and classical MPC.
It turns out that most of the best-performing ORL algorithms are not much different in performance from model-free online RL algorithms.
They are all better than vanilla imitation learning (behavioral cloning), but there is a certain gap with the MPC method in terms of task response accuracy.
Our main conclusion is that the generalization ability of the current ORL policies is limited in the absence of dynamic model information for legged locomotion tasks. 
Furthermore, it addresses the importance on how to enhance the rapid adaptability of ORL policies to accommodate environmental stochasticity and non-stationarity.
We will make the dataset public to facilitate ORL algorithms research towards realistic applications.

\section{Related Work}
\textbf{ORL benchmark:} 
Recent works have proposed benchmarks for ORL algorithms to standardize community and research platforms.
\cite{DBLP:conf/icml/AgarwalS020} released a challenging and well-known discrete DRL benchmark for the Atari 2600 games.
\cite{DBLP:journals/corr/abs-1910-01708} also generated datasets on Atari games using a single partially-trained behavioral policy, and benchmarked the performance of several ORL algorithms.
RL Unplugged \cite{gulccehre2020rl} contained not only datasets from Atari games, but also DM control suites. 
Furthermore, \cite{DBLP:journals/corr/abs-2004-07219} released datasets (D4RL) including some simulation environment tasks, which aim to highlight challenging properties that may exist in real-world applications.  
\cite{DBLP:journals/corr/abs-2102-00714} proposed a near real-world ORL benchmark with datasets from simulated and realistic environments, and additional test datasets for policy validation.
Recently, \cite{gurtler2023benchmarking} presented a benchmark consisting of simulated and real data performed on a dexterous manipulation platform with 9 degrees of freedom.
Compared with these works, our dataset contains realistic quadrupedal locomotion tasks with 12 degrees of freedom, which is more challenging for ORL.


\textbf{Robotic learning benchmark:} There are also a large body of works establishing benchmarks for robotic learning.
Several works have established DRL benchmarks for robotic manipulation tasks, and evaluated model-free DRL algorithms. 
Some simple tasks on legged \cite{DBLP:conf/corl/AhnZHP0LK19} and mobile \cite{DBLP:conf/corl/MahmoodKVMB18} robots were also considered.
\cite{DBLP:journals/corr/abs-1905-07447} proposed a reproducible, low-cost arm benchmark platform for robotic learning and present a small-scale realistic robotic grasping dataset. 
Moreover, \cite{DBLP:conf/corl/YuQHJHFL19} presented a benchmark consisting of 50 different robotic manipulation tasks for evaluation of meta-reinforcement learning and multi-task learning. 
\cite{DBLP:conf/nips/WolczykZPKM21} further set new tasks on this dataset, establishing a robotics benchmark for continuous reinforcement learning.
These works are mainly aimed at robot manipulation tasks or establish model-free DRL simulation platforms and evaluation benchmarks.
Our work is to construct the realistic quadrupedal datasets and benchmark for ORL research.
\section{Preliminary}
In the standard reinforcement learning setting, we model the interaction between the agent and the environment as a Markov Decision Process (MDP) \cite{sutton2018reinforcement}, denoted by the tuple $(\mathcal{S}, \mathcal{A}, P, R, \rho_0, \gamma)$, where $\mathcal{S}$ and $\mathcal{A}$ denote the state and action spaces, $P(\bs'|\bs,\ba)$ is the stochastic transition dynamics, $R(\bs,\ba)$ is the reward function,  $\rho_0(\bs)$ is the initial state distribution, and $\gamma \in [0,1]$ is the discount factor. 
The aim is to find a policy that maximizes the expected discounted rewards $J(\pi) := \mathbb{E}_{\tau \sim \pi}\left[ \sum_{t=0}^{\infty} \gamma^t R(\bs_t, \ba_t) \right]$ in the environment, where $\tau = \left( \bs_0, \ba_0, R_0, \bs_1, \ba_1, \dots \right)$ is a sample trajectory. 
We utilize $\tau \sim \pi$ to indicate that the trajectory distribution $p(\tau)$ depends on the policy $\pi$, where $\bs_0 \sim \rho_0(\bs)$, $\ba_t \sim \pi(\ba_t | \bs_t)$, and $\bs_{t+1} \sim P(\bs_{t+1} | \bs_t, \ba_t)$. 
We also define the action-value function $Q^\pi(\bs, \ba) := \mathbb{E}_{\tau \sim \pi}\left[ \sum_{t=0}^{\infty} \gamma^t R(\bs_t, \ba_t) \vert \bs_0 = \bs, \ba_0 = \ba \right]$, which describes the expected discounted rewards starting from state $\bs$ and action $\ba$ and following $\pi$ afterwards, and the state value function $V^\pi(\bs)=\mathbb{E}_{\ba \sim \pi(\ba | \bs)}\left[ Q^\pi(\bs, \ba) \right]$. 

In ORL, the agent is provided with a static data $\mathcal{D} = \left\{ \tau \right\}$ which consists of  trajectories collected by running behavior polices.
Therefore, the data distribution in the dataset will directly affect the performance of the ORL policies. 
Behavioral policy currently used to collect data include trained online RL policies, control methods, and human demonstrations.
For complex quadrupedal robot locomotion tasks, it is also important to design reasonable and reliable metrics to evaluate the performance of ORL policies.

\section{Realistic Quadrupedal Benchmark}
We construct a novel ORL benchmark based on a real-world quadrupedal locomotion dataset and aim to evaluate the realistic performance of ORL algorithms. 
In this section, we first present the evaluation metrics, which are utilized for the evaluation and testing of ORL algorithms.
Then, we introduce the design for real-world tasks, including the environmental terrain and locomotion commands. 
Next, we discuss the properties of our proposed benchmark, and elaborate on the settings of the basic elements of the MDP contained in the dataset. 
Finally, we discuss limitations of proposed benchmark.

\subsection{Metrics}
Establishing reasonable metrics is crucial for ORL and robot learning research. 
In the proposed benchmark, we mainly design metrics from three aspects: return, energy consumption and policy stability.

For online policy evaluation, we deploy the trained ORL algorithms to the realistic robot, and calculate the cumulative undiscounted reward 
(\textbf{Return}) 
$
\mathbb{M}_1 = \sum_t^{T} r_t,
$
where $T$ is the number of real-world interactions.
This metric mainly reflects the robot's task response accuracy.
It will directly serve as a measure of the final performance of the algorithm, which is a common metric in the DRL community.

Furthermore, the dimensionless cost of transportation (\textbf{COT}) is a metric in the legged locomotion research, and is defined as
$
\mathbb{M}_2 = \sum_t^{T}[(|\tau_t\dot{q_t}|)/(mg\Vert v_t \Vert_2)]/T,
$
where $mg$ and $v$ are the total weight and linear velocity of the robot, respectively.
Intuitively, this quantity represents the positive mechanical power applied by the actuator per unit weight and unit locomotion speed \cite{collins2005efficient}.
We utilize the COT metric to compare the energy consumption of ORL policies on different real-world tasks.

To quantitatively reflect the instability of ORL algorithms in realistic robot, the coefficient of variation (\textbf{COV}) is utilized as a novel evaluation metric, which is a dimensionless statistic to measure the degree of data dispersion
$
\mathbb{M}_3 = \sigma(\mathbb{M}_1^i)/\mu(\mathbb{M}_1^i),
$
where 
$\mu(\mathbb{M}_1^i)$ and $\sigma(\mathbb{M}_1^i)$ are 
the mean and standard deviation of the return for the $i^{th}$ task, respectively.
The COV metric measures the stability of an ORL algorithm, which is specifically reflected in the fluctuations of Return. 
The larger the COV value is, the more unstable the algorithm is, and the easier it is to be disturbed by external factors.

\subsection{Tasks design}
Compared to simulation-based datasets, our datasets can more accurately reflect the challenges faced by ORL algorithms when dealing with real-world problems.
The dataset contains five environments and six locomotion commands, which means that the dataset contains a total of thirty tasks.

In the initial release of the dataset, five common terrains are considered, including indoor floor, asphalt road, marble slope, marble floor and grassland.
These terrains are represented by the mathematical symbol $M$ in order, and their values range from $1\ \sim \ 5$.
For specific environmental parameters, we consider friction coefficient and slope. 
We utilize a flat ground with a marble surface as our base environment, as it provides common features for quadruped robot locomotion tasks, like moderate friction ($0.4\ \sim \ 0.6$).
Then we further expand the complexity of the environment, such as increasing the slope of the terrain or friction coefficient. 
The ability of a robot to adapt to complex terrain depends on its mechanical structure and control policy performance, so all differences between environments must be within a controllable range.

We also provide locomotion diversity in the dataset by designing multiple desired linear and angular velocities for the robot's center of mass (COM).
We set up six sets of locomotion commands in each environment, which are $(0.4\, m/s, 0)$, $(0.2\, m/s,0)$, $(0.4\, m/s, 0.5\, rad/s)$, $(0.2\, m/s, 0.5\, rad/s)$, $(0.4\, m/s, -0.5\, rad/s)$ and $(0.2\, m/s, -0.5\, rad/s)$.
These commands are represented by the mathematical symbol $N$ in order, and their value range is $1\ \sim \ 6$.

\begin{figure}[t]
\includegraphics[width=0.5\textwidth,height=0.3\textwidth]{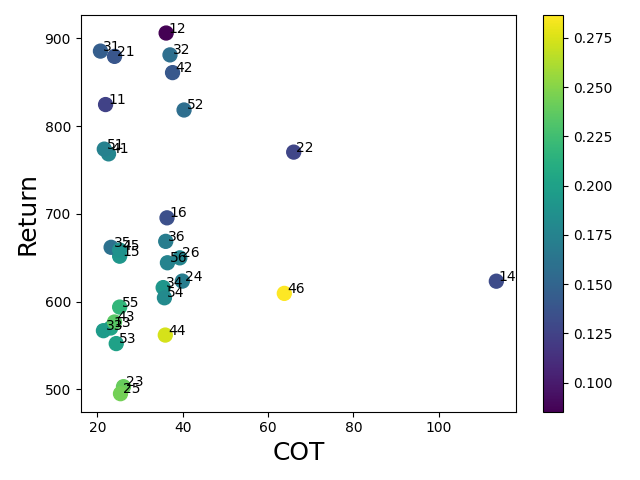}
\caption{
Data distribution on the Return, COT and COV metrics. 
The yellower the color of the discrete point, the larger the COV value, otherwise vice versa.
The number indicates task number $MN$.
} 
  \label{fig:distribution}
\end{figure}
The online model-based MPC \cite{DBLP:conf/iros/CarloWKBK18} method was leveraged to collect robots' empirical data.
MPC builds a simplified robot dynamics model and solves convex optimization problems in real-time. 
The MPC has stable and effective control over standing, trotting, bouncing, and other behaviors.
Figure \ref{fig:distribution} show that the diversity of data distribution collected using MPC is reflected in return, stability and energy consumption. 
The Return for most tasks are in the range of 500 to 900, indicating the distribution of data for "Medium" and "Expert". 
The COT of most tasks range from 20 to 40. 
Moreover, all tasks' COV are between 0 and 0.5.

The diversity of dataset primarily comes from two sources: the different controllers and the environmental noise.
For the MPC controllers setup, parameters had been adjusted according to each task, such as desired pitch angles, foot position heights, and model prediction horizons $T$. 
Specifically, we leveraged the well-tuned MPC with prediction horizon $T$ ($8\sim 12$) to collect "Expert" data and tuned the prediction horizon down to $T/2$ ($4\sim 6$) to collect "Medium" data.
In effect, our "Expert" data is "Near-Expert" due to environmental noise in nature.
Specifically, physical characteristics such as the coefficient of friction and slope are not always constant. 
Temperature change and parts abrasion also affect the actuation of the robot. 
These environmental noises vary in each run, resulting in performances varying.
So there are also differences among even the same "Medium" or "Medium-Expert" type tasks.


\subsection{Characteristics of proposed benchmark}
Our benchmark suite is designed to cover properties possible in real-world applications to determine the difficulty of learning problems and the choice of policies. 
The properties of the proposed benchmark are discussed and summarized.
\textbf{Stochasticity}:
One of the main challenges of real-world tasks is the stochasticity of the realistic environment.
There are many sources of stochasticity, such as the ground friction, ground hardness, robot motor damping and motor friction.
An unstable ORL policy will cause the quadruped robot to deviate from the desired locomotion command. 
These stochasticity are unavoidable and difficult to control, so higher requirements are placed on the robustness and stability of the policy.
\textbf{Scarcity}: 
Previous benchmarks were mostly built in simulated environments. 
For real-world robot locomotion tasks, there is no fairly realistic simulation platform that can accurately describe the stochasticity in reality. 
Real-world data collection is time-consuming and labor-intensive, making it difficult to obtain a large amount of real-world data.
This requires the ORL algorithm to have better sample efficiency.
\textbf{Conservativeness}: 
Random exploration of robots in real-world scenarios brings additional hardware costs and potential risks.
In our benchmark marble slope task, once the robot takes aggressive actions, the stride is easy to be too large and the center of gravity is unstable. 
Non-conservative behavior always puts the robot at risk and wears out the hardware, so real-world policies often act conservatively.
This will result in a real-world dataset that is hardly more diverse than the benchmarks in the simulated environment.

\subsection{Actions, observations and rewards design}
For the design of the action space, in the DRL-based quadrupedal locomotion field, most researchers \cite{RoboImitationPeng20} consider utilizing the desired joint angles (12 dims), and then calculate the torque through a PD controller $\tau=K_{p}(\hat{q}-q)+K_{d}(\hat{\dot{q}}-\dot{q})$ to control the locomotion of the quadruped robot, where $q$ and $\dot{q}$ are the current joint angle and angular velocity, $\hat{q}$ and $\hat{\dot{q}}$ are the desired joint angle and angular velocity, $K_{p},K_{d}$ and $\tau$ are the proportional parameter, differential parameter and desired torque.
\cite{DBLP:conf/sca/PengP17} experimentally demonstrate that policies with such action space can achieve better performance.
So in the proposed benchmark, we utilize the target joint angle as the action space.
The state space design mainly considers the proprioception of the robot, which consists of a 44-dimensional continuous vector. 
Specifically, this vector contains COM linear velocity (2 dims), attitude angle (3 dims) and angular velocity (3 dims), joint angle (12 dims) and joint angular velocity (12 dims), action at the last time step ( 12 dims).
Such state design is an common choice in the DRL-based quadrupedal locomotion studies. 
When designing the reward function, some previous DRL-based quadrupedal locomotion research \cite{eth2020} considered rather complex factors. 
This usually requires a great deal of expert knowledge as well as hyperparameter tuning time.
In contrast, we mainly consider the robot's responsiveness to the locomotion command and energy consumption: $r=r_1+r_2+0.001*r_3$, where
$r_1 = e^{-\sum(\hat{v}-v)^2/0.025}$,
$r_2 = e^{-\sum(\hat{\theta}-\theta)^2/0.025}$, and 
$r_3 = -\sum\tau^2/12$.
$\hat{v},v,\hat{\omega},\omega, \hat{\theta}, \theta$ and $\tau$ represent the desired linear velocity, current linear velocity, desired yaw rate, yaw rate, desired pitch angle, current pitch angle, and desired torque, respectively.

\subsection{Limitations}
Two main limitations of the proposed benchmark are summarized.
\textbf{Observation}: 
We currently only utilize the robot's proprioception, which is the key to achieve fast computing requirements under the robot's limited computing resources.
However, richer inputs can enable robots to complete more complex tasks, such as visual information that can help robots avoid obstacles or complete navigation tasks.
\textbf{Task complexity}: 
During the collection of the dataset, the locomotion control of the \textit{Unitree A1} robot relies heavily on its control policies. 
The environment and tasks in the current dataset are relatively common. 
These scenarios are the basic composition of the dataset, and the ability of the robot to respond to challenging terrains requires higher performance policies. 
Including richer and more diverse data is an important direction for future work. 
More rough terrain, flexible and diverse robotic behavioral skills, and exterior sensor information will be considered.

\section{Methods}
\label{Methods}
In this section, Three baselines are first introduced, and then we summarize several advanced ORL algorithms that will be trained and tested on the proposed benchmark.

\textbf{Baselines:}
\textbf{Behavioral cloning} (\textbf{BC}) \cite{DBLP:conf/nips/Pomerleau88} is a simple imitation learning method that directly trains policies to imitate the behavioral policies in the dataset without any reward function. This method serves as a weak baseline for ORL algorithms. 
\textbf{MPC} \cite{DBLP:conf/iros/CarloWKBK18} for the collected data will serve as a strong baseline for the ORL algorithms because of its high performance and stable control.
Moreover, \textbf{DreamWaQ} \cite{nahrendra2023dreamwaq}, an algorithm with only proprioception to achieve state-of-the-art performance on \textit{Unitree A1} robot, will serve as an model-free online RL baseline. 
The core idea of the algorithm is to utilize implicit terrain imagination and environmental representation for potential evaluation and inference.

\textbf{ORL algorithms:}
Since ORL algorithms are learned from a pre-collected dataset, directly exploiting off-policy algorithms will suffer from extrapolation error or distribution shift, where the training policy tries to reach out-of-distribution states and actions. 
Therefore, model-free ORL usually restrict the learned policy explicitly or implicitly to be close to the behavioral policy in offline data.
\textbf{Direct policy constraints:}
\textbf{BCQ} \cite{DBLP:conf/icml/FujimotoMP19} is the first practical ORL algorithm that directly restricts the policy, and its policy function is expressed as a combination of a conditional VAE and a perturbation function to compensate for extrapolation errors in target value estimation.
\textbf{Implicit policy constraints:}
\textbf{BEAR} \cite{DBLP:conf/nips/KumarFSTL19} is a SAC-based ORL that limits the support of policy functions in the data distribution by minimizing the maximum mean difference between the policy function and the approximate behavioral policy function.
\textbf{PLAS} \cite{DBLP:conf/corl/ZhouBH20} is an ORL algorithm whose policy function is trained in the latent space of a conditional VAE and uses fewer constraints.
\textbf{TD3PlusBC} \cite{DBLP:conf/nips/FujimotoG21} is a TD3-based ORL algorithm, which mainly introduces a regular term similar to the BC method when updating the policy.
\textbf{AWAC} \cite{DBLP:journals/corr/abs-2006-09359} is also a TD3-based ORL algorithm that uses the Q-function to estimate the odds function to reduce its variance and improve sample efficiency. 
\textbf{CRR} \cite{wang2020critic} is similar to AWAC in that it actively filters out below-average actions using an indicator function, but chooses a more pessimistic advantage estimator.
\textbf{IQL} \cite{DBLP:journals/corr/abs-2110-06169} can avoid querying values for unseen actions, while being able to perform multi-step dynamic programming updates.
\textbf{Regularization:}
\textbf{CQL} \cite{DBLP:conf/nips/KumarZTL20} is a SAC-based ORL that mitigates overestimation errors by minimizing the action value under the current policy and maximizing the value under the data distribution of the underestimation problem.
\textbf{Model-based methods:}
\textbf{MOPO} \cite{DBLP:conf/nips/YuTYEZLFM20} is a model-based ORL algorithm for offline policy optimization, which utilizes a probabilistic ensemble model to generate new data with an uncertainty penalty.
\textbf{COMBO} \cite{DBLP:conf/nips/YuKRRLF21} is similar to MOPO, but it also exploits the conservative loss proposed in CQL.

\section{Experiments}
In this section, the performance of 11 ORL algorithms is first evaluated, followed by a stability analysis, and finally a comparative experiment with the baseline algorithms.
Statistics on all tasks and more algorithm comparison experiments are provided in the supplementary video.
\subsection{Real-world performance}
\begin{figure}[t]
  \centering
\includegraphics[width=0.4\textwidth,height=0.28\textwidth]{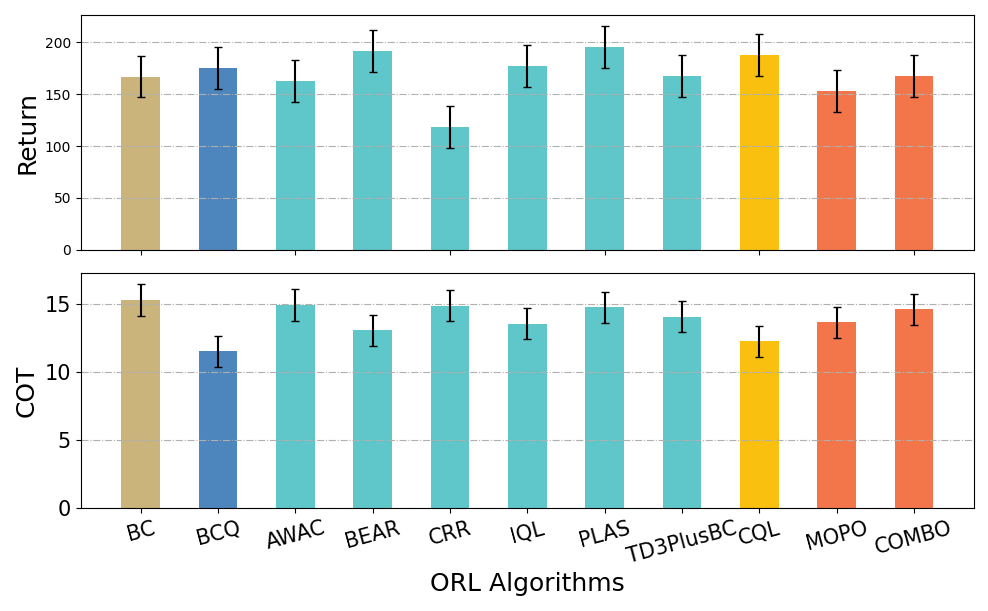}
  \caption{
  Performance comparison of ORL algorithms on Return and COT metrics across all realistic tasks.
The black vertical line indicates one standard deviation. 
Three random seeds are used.
  }
  \label{fig:ope_return_cot}
\end{figure}
\begin{figure}[t]
  \centering
\includegraphics[width=0.4\textwidth,height=0.25\textwidth]{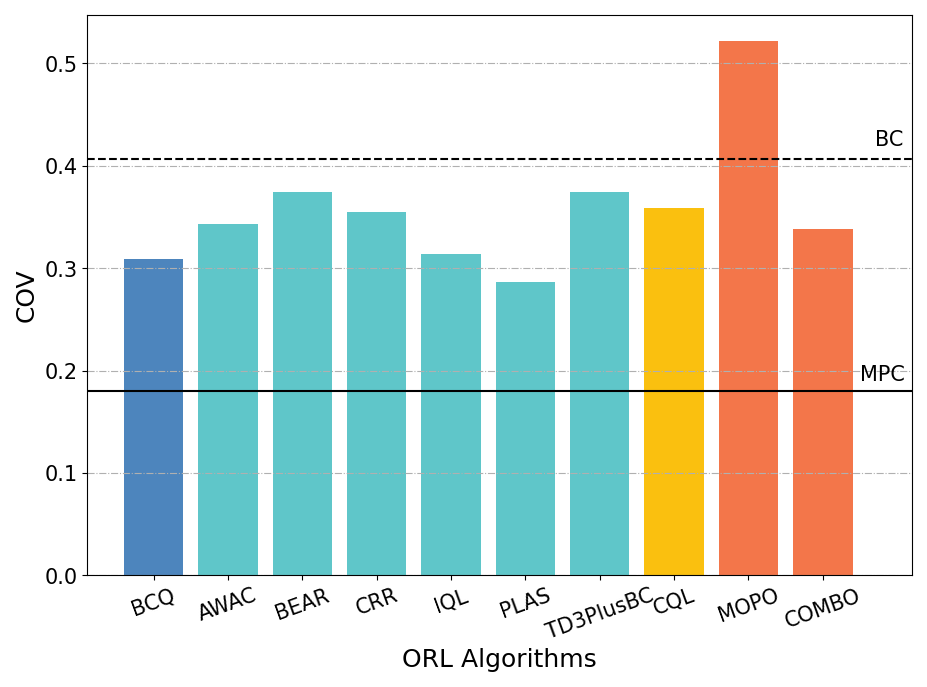}
  \caption{
  Instability of the ORL algorithm's return on different random seeds for all tasks.
  The black dashed and solid lines indicate the COV of the BC and MPC algorithms, respectively.
  }
  \label{fig:ope_cov}
\end{figure}
\begin{figure}[tbp]
  \centering
   \subfigure{
\includegraphics[width=0.46\textwidth,height=0.3\textwidth]{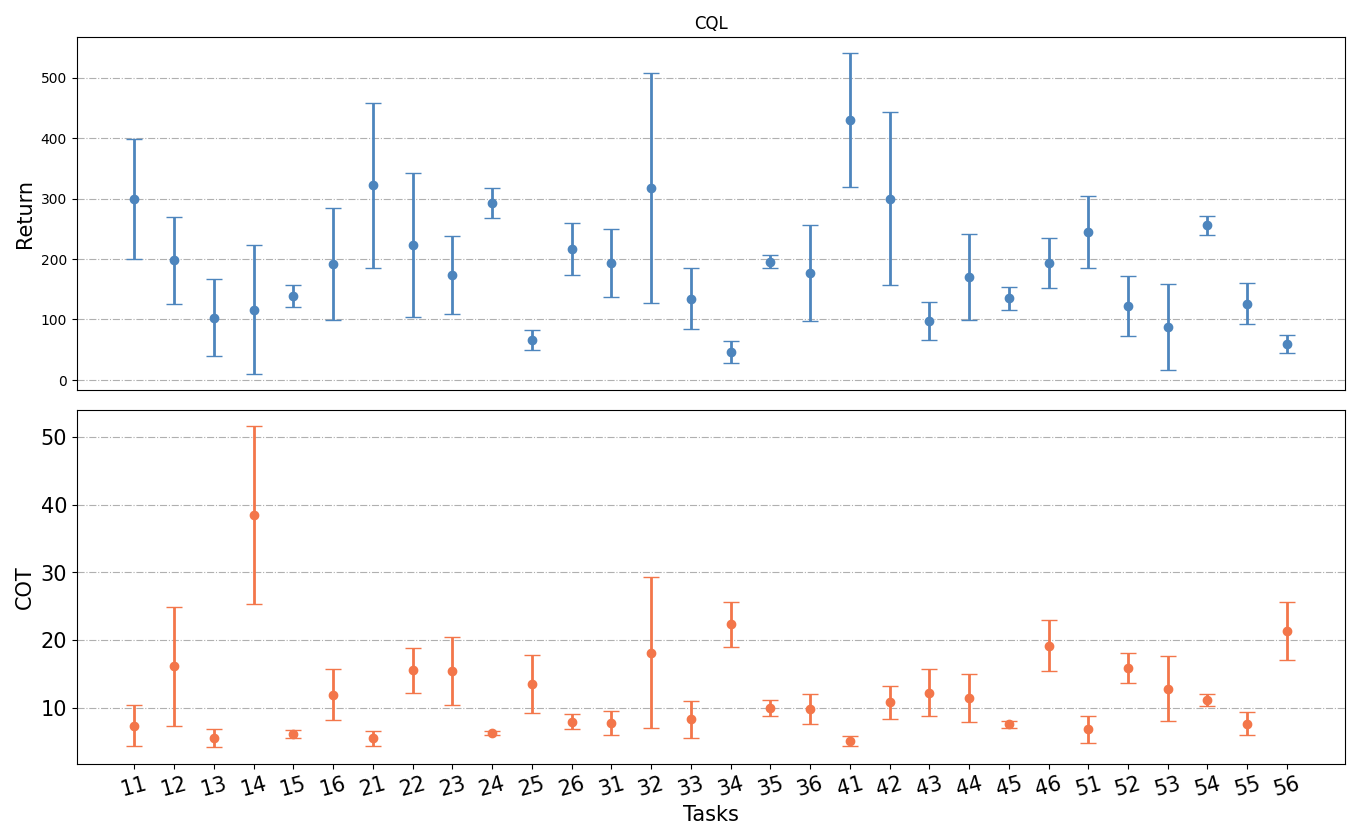}
}
  \caption{
Fluctuations in the return and energy consumption of CQL algorithm across all real-world environments and tasks.
The horizontal axis number indicates the task number $MN$.
The dots and vertical bars represent the mean and one standard deviation, respectively.
 Three random seeds are used.
  }
  \label{fig:cql_return_cot}
\end{figure}

In Figure \ref{fig:ope_return_cot}, we compare the average return and energy consumption of ORL algorithms on all real-world tasks by utilizing Return and COT metrics.
Compared with other ORL algorithms, CQL and BCQ achieve better performance on both metrics. 
This is in line with the results reported in the common D4RL \cite{DBLP:journals/corr/abs-2004-07219} dataset where the hyperparameter tuned CQL has the best performance, followed by BCQ.
The Return metric of PLAS, BEAR, and IQL is high, but the COT metric is too large, indicating that these algorithms consume more energy in exchange for better return. 
Moreover, the performance of TD3PlusBC and COMBO is mediocre, slightly better than the baseline BC, while the performance of AWAC and MOPO is a little worse than that of BC. 
CRR is the worst-performing algorithm, with the lowest return and slightly less energy consumption than that of BC.
The recent benchmark \cite{gurtler2023benchmarking} on realistic manipulation tasks shows that CRR performs best on real-world dexterous manipulation tasks, while CQL performs the worst. 
This does not seem to be consistent with our experimental results.
We suspect that the reason may be that the task settings are inconsistent and the algorithms have different preferences. 
Another possible cause is the influence of algorithmic hyperparameters.
\begin{figure*}[tbp]
  \centering
   \subfigure{
\includegraphics[width=0.4\textwidth,height=0.25\textwidth]{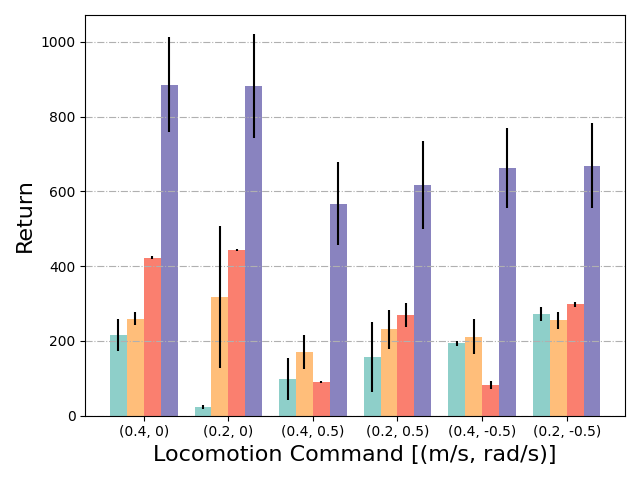}
}
  \subfigure{
\includegraphics[width=0.4\textwidth,height=0.25\textwidth]{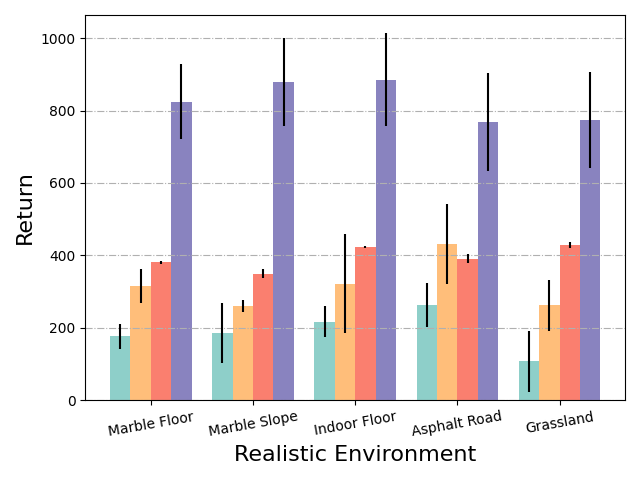}
}
  \subfigure{
\includegraphics[width=0.6\textwidth,height=0.02\textwidth]{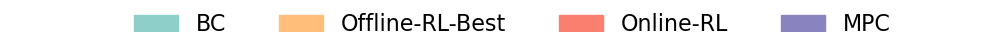}
}
  \caption{
Performance comparison of different tasks (left) in the Indoor floor and performance comparison of different environments (right) with locomotion command (0.4 m/s, 0 rad/s).
     "Online-RL" indicates the \textbf{DreamWaQ} algorithm.
    "Offline-RL-Best" indicates the ORL algorithm with the best performance under the corresponding task (environment). 
    They are PLAS, CQL, AWAC, PLAS, AWAC and BCQ algorithms in different tasks (left), and AWAC, CQL, PLAS, CQL and BEAR algorithms in different environments (right). 
    The black vertical line indicates one standard deviation. 
    Three random seeds are used.
  }
  \label{fig:comparison_online_rl}
\end{figure*}
\begin{figure}[tbp]
  \centering
  \subfigure{
\includegraphics[width=0.46\textwidth,height=0.3\textwidth]{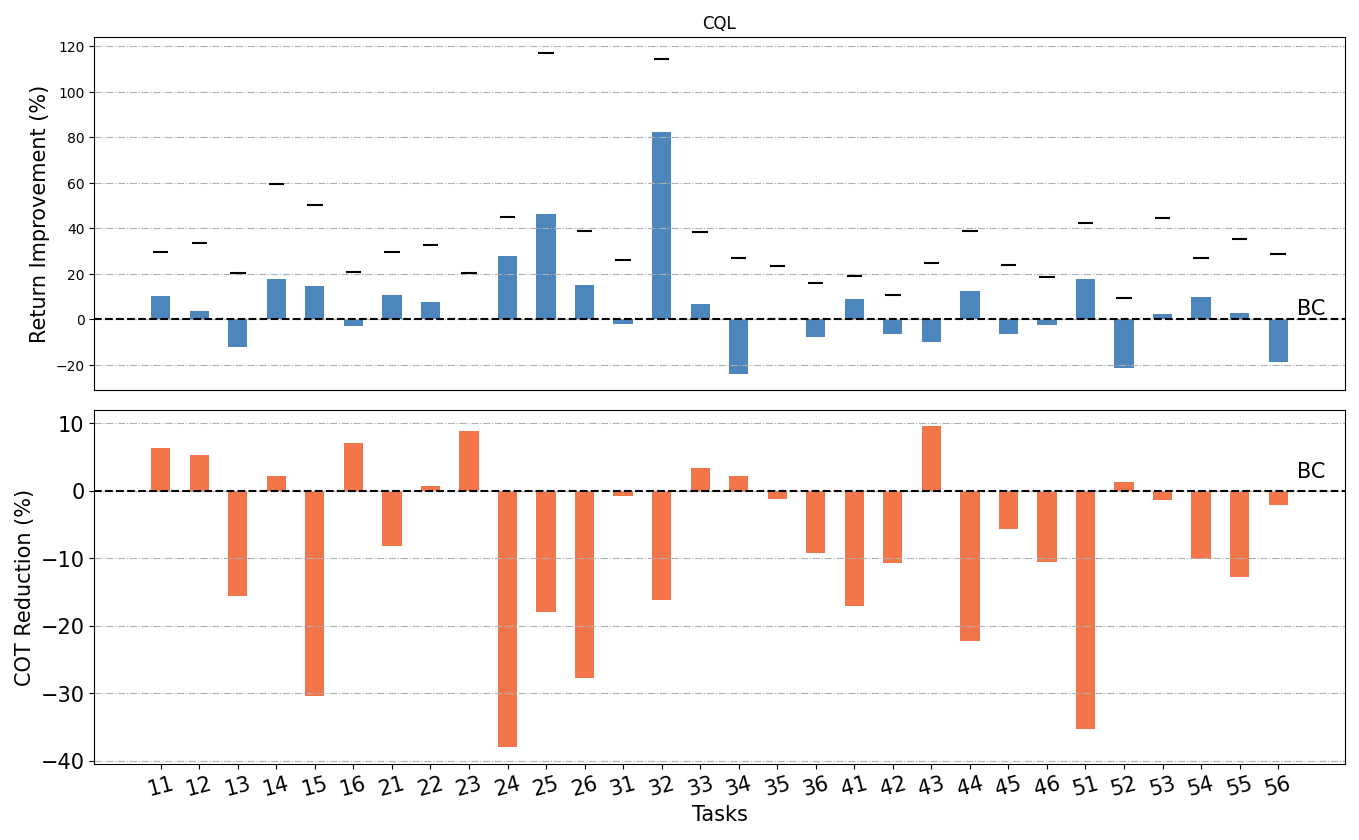}
}
  \caption{
  The return improvement of CQL (upper) and the energy consumption reduction (bottom). We compare the return improvement and energy consumption reduction for the CQL algorithm compared to the BC algorithm. 
  The horizontal axis number indicates the task number $MN$.
  The black dashed and black solid lines represent the performance of the baseline BC algorithm and MPC, respectively.
  }
  \label{fig:cql_return_imp}
\end{figure}

\subsection{Stability analysis}
We utilize the COV metric to quantitatively characterize the instability of ORL algorithms on realistic robot, as shown in Figure \ref{fig:ope_cov}. 
Most ORL algorithms show improvement over the baseline BC, except for the MOPO algorithm. 
PLAS, IQL and BCQ have lower COV, which means better stability during actual operation. 
However, the COV of all ORL algorithms is higher than that of the MPC. 
The results show that the stability of the ORL algorithms is not as good as that of MPC, and there is much room for improvement.

We also plot the fluctuations of the Return and COT of the CQL algorithm across all real-world environments and tasks (Figure \ref{fig:cql_return_cot}).
Specifically, the CQL performs well on several tasks, but the Return and COT metrics on tasks with task numbers such as 14 and 32 fluctuate greatly, indicating that the CQL may someties struggle to run stably. 
In fact, the current version of the dataset contains not much difference among tasks or environments. 
Such phenomenon shows that the policies trained by the same algorithm on different data distributions have different degrees of generalization, that is, there may be underfitting and overfitting.


\subsection{Comparison with baselines}
To investigate the performance differences among different types of algorithms, we set up comparative experiments with different tasks or environments, as shown in Figure \ref{fig:comparison_online_rl}. 
The MPC for data acquisition has the best performance, and the vanilla imitation learning BC has the worst performance. 
The best ORL algorithm is not much different from the model-free online RL algorithm in performance, and even performs higher in some tasks and environments.

MPC is superior to other algorithms mainly because it is an online algorithm that contains information about the robot dynamics models and can be optimized in real time.
Such capability is essential for legged locomotion control, as the environment and tasks may change over time.
For learning-based algorithms (BC, online RL and offline RL), the new environments and tasks faced in the real robot deployment phase mainly rely on the generalization ability of the trained policy. 
However, such generalization ability is limited in the absence of dynamic model information, and cannot be compared with classical MPC methods in terms of task response accuracy.
We conclude that learning-based policies need to take into account dynamic model information and rapid adaptability to better respond to real-world tasks.

To further investigate the generalization ability and stability of the ORL policies on our dataset, we compare the performance of the ORL and BC algorithms.
The return improvement and the energy consumption reduction for CQL algorithm are reported in Figure \ref{fig:cql_return_imp}.
Figure \ref{fig:cql_return_imp} shows that compared with the BC algorithm, CQL's return has a certain improvement in several real-world tasks (tasks numbered 25 and 32) and the drop in energy consumption (tasks numbered 24 and 51). 
CQL's performance is sometimes much lower than that of the MPC. 

In summary, the dataset collected by classical MPC is rather challenging for ORL, and our dataset has higher requirements for the generalization and adaptability of ORL policy.
If the performance of the ORL policy trained on our benchmark is comparable to that of MPC, it means that such policy is mature enough to be more widely utilized in reality. 
This is also exactly why MPC is utilized to collect data in our dataset instead of online RL.

\section{Conclusions}
We present a benchmark of challenging quadrupedal robot locomotion designed to help improve the latest techniques for ORL research.
We utilize a relatively mature MPC method to collect datasets instead of using the online RL algorithm.
Analysis and evaluation of all tasks show that the ORL algorithms has a lot of room for improvement on real robot tasks.
Current ORL (or even learning-based) algorithms are inferior to MPC method in terms of task response and stability control.
Practical application-oriented ORL research needs to consider more goals and constraints, such as robot dynamics model information, and rapid adaptation.
We hope that the benchmark can shed some light on future research and draw more attention to realistic ORL-based applications.


\clearpage
\bibliographystyle{ieeetr}
\bibliography{IEEEexample}

\end{document}